\documentclass[conference]{IEEEtran}
\IEEEoverridecommandlockouts
\usepackage{cite}
\usepackage{amsmath,amssymb,amsfonts}
\usepackage{algorithmic}
\usepackage{graphicx}
\usepackage{textcomp}
\usepackage{xcolor}
\usepackage{multirow} 
\usepackage{makecell}
\usepackage{hyperref}
\def\BibTeX{{\rm B\kern-.05em{\sc i\kern-.025em b}\kern-.08em
    T\kern-.1667em\lower.7ex\hbox{E}\kern-.125emX}}
    
\makeatletter
\def\ps@IEEEtitlepagestyle{%
\def\@oddfoot{\mycopyrightnotice}%
\def\@evenfoot{}%
}
\def\mycopyrightnotice{%
\gdef\mycopyrightnotice{}
}
\begin{document}

\title{Bit-depth enhancement detection\\for compressed video\\
}

\author{
\IEEEauthorblockN{Nickolay Safonov\IEEEauthorrefmark{1}, Dmitriy Vatolin\IEEEauthorrefmark{1}\IEEEauthorrefmark{2}}
\IEEEauthorblockA{\IEEEauthorrefmark{1}Lomonosov Moscow State University, Moscow, Russia \\ \IEEEauthorrefmark{2}MSU Institute of Advanced Studies of Artificial Intelligence and Intelligent Systems, Moscow, Russia \\  {\tt\small \{nikolay.safonov, dmitriy\}@graphics.cs.msu.ru}\\}
}


\maketitle

\begin{abstract}
In recent years, display intensity and contrast have increased considerably. Many displays support high dynamic range (HDR) and 10-bit color depth. Since high bit-depth is an emerging technology, video content is still largely shot and transmitted with a bit depth of 8 bits or less per color component. Insufficient bit-depths produce distortions called false contours or banding, and they are visible on high contrast screens. To deal with such distortions, researchers have proposed algorithms for bit-depth enhancement (dequantization). Such techniques convert videos with low bit-depth (LBD) to videos with high bit-depth (HBD). The quality of converted LBD video, however, is usually lower than that of the original HBD video, and many consumers prefer to keep the original HBD versions. In this paper, we propose an algorithm to determine whether a video has undergone conversion before compression. This problem is complex; it involves detecting outcomes of different dequantization algorithms in the presence of compression that strongly affects the least-significant bits (LSBs) in the video frames. Our algorithm can detect bit-depth enhancement and demonstrates good generalization capability, as it is able to determine whether a video has undergone processing by dequantization algorithms absent from the training dataset. 
\end{abstract}


\section{Introduction}
Traditionally, storing a video pixel’s color component employs 8 bits. Using 256 gradations of each color was sufficient for early digital displays that had a maximum brightness of about 100 nits: the transitions between nearby colors looked smooth. In recent years, however, more and more devices enable high dynamic range and have screens with higher contrast. Foremost are HDR (High Dynamic Range) TV screens, although many smartphones also support this feature. On such screens, distortions caused by quantization of scene light into discrete values become visible: changes between nearby colors are noticeable, resulting in artificial bands in the image. This unwanted effect is called banding or false contours. 

\begin{figure}[tbp]
\centerline{\includegraphics[width=0.48\textwidth]{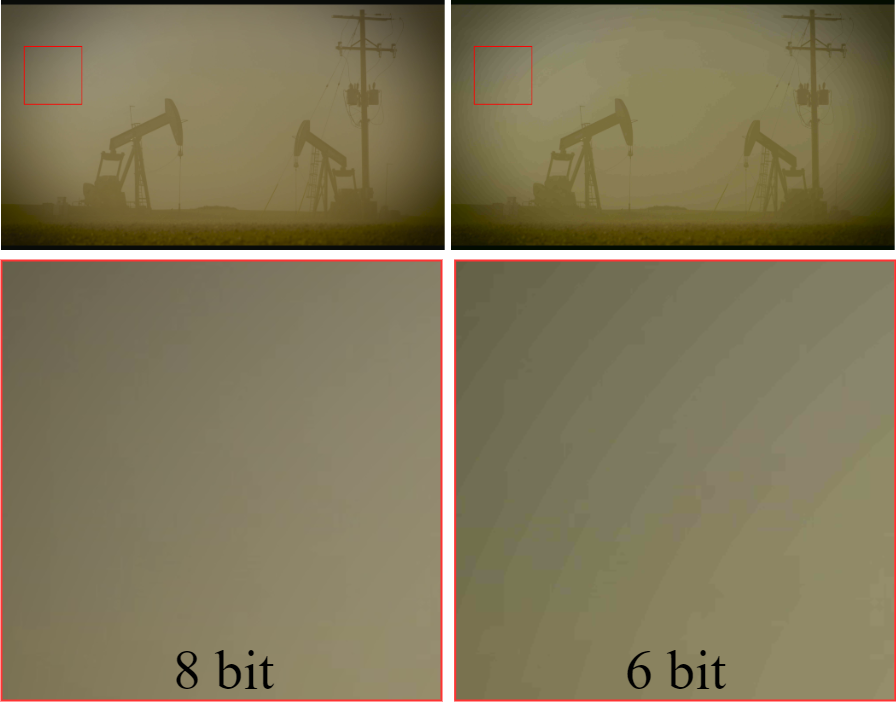}}
\caption{Illustration of insignificant bit-depth artifacts: false contours and color distortions.}
\label{fig:distortion}
\end{figure}

Formats with high bit-depths are becoming more widespread. In the high-quality-television industry, the 10-bit format is practically a standard. Formats that support 12 bits are also under development~\cite{recommendation2012parameter} and serve in the entertainment industry. Additional bits allow expansion of the transmitted signal’s color gamut (for example, HDR and WCG, or wide color gamut) as well as reduction of the quantization step, thereby smoothing the gradients between nearby color values to eliminate distortions resulting from insufficient bit-depth. Figure \ref{fig:distortion} demonstrates such distortions.

Researchers developed band-masking algorithms (debanding algorithms) and LSB-recovery (bit-depth-enhancement) algorithms to convert low bit-depth (LBD) video or images to high bit-depth (HBD) video or images (i.e., to perform so-called up-conversion). Debanding methods take an image as input and attempt to smooth or redistribute pixel values in areas with noticeable distortions. They usually detect areas with banding and only operate on those areas. The reconstruction algorithms try to predict LSBs in an HBD video. 

The emergence of neural networks has caused rapid proliferation of bit-depth enhancement algorithms; some examples noticeably improve the visual quality of images. Certain algorithms perform bit-depth enhancement through per-frame restoration or by aggregating information from several video frames, thereby increasing the recovery quality.

Up-converted images and videos have worse visual quality than true-HBD images and videos because the algorithms are imperfect. True high bit-depth means the source video was captured and stored in that format. Many consumers prefer original HBD videos over converted ones, so a relevant problem is detection of HBD videos created by up-conversion from LBD videos. 

Several algorithms can detect and assess banding visibility. One disadvantage, however, is that the noticeability of banding can vary with contrast and resolution; the abovementioned methods target standard displays and require nontrivial parameter tuning to work with screens that have a strong contrast ratio. No current banding-detection method directly predicts the video’s true color depth, a characteristic that affects video quality. Bit-depth strongly affects visual quality, and true HBD videos have higher quality than up-converted ones~\cite{Bot2021ASSESSMENTOQ}. But HBD and up-converted LBD are often indistinguishable by objective-quality metrics, including those based on banding-detection algorithms.

Our paper’s main contribution is a new method for detecting up-conversion that took place after video compression, which hides manipulation of LSBs. We aim to detect up-conversion from 6 to 8 bits and from 8 to 10 bits — the most widely used bit-depths. The technique generalizes well even to dequantization algorithms whose outputs are absent from the training set. The code and models are publicly available at \url{https://github.com/msu-video-group/BDEDM}.

\section{Related work}
To the best of our knowledge, no published work aims to detect bit-depth enhancement (up-conversion). Since most video is distributed in a compressed state, our aim is detection of bit-depth increases in compressed video, but this last condition complicates the problem. Bit-depth enhancement algorithms are strongly related to the subject, so we review them briefly. 

The following are the two most common approaches to suppressing distortions that result from insufficient video bit-depth:
\begin{itemize}
\item Image filtering or pixel-value redistribution by adding noise (dithering). 
\item Attempting to restore the image’s LSBs.
\end{itemize}

The first approach may not increase the video’s bit-depth , but when applied to, for example, a 10-bit image with 8 high bits corresponding to some 8-bit image and 2 low bits corresponding to noise or zeros, the output is an approximation of the LSBs, and the distortion becomes slightly less noticeable. The first step in this type of approach is usually to locate regions with visible banding; the second is to smooth these areas and thereby suppress the artifact’s visibility. In~\cite{bhagavathy2009multiscale}, for example, the authors analyze the image at multiple resolutions to detect banding and then make the banding less visible through probabilistic dithering. The authors of~\cite{baugh2014advanced} segment the input images by color-channel values to identify large single-color regions and, afterward, also apply dithering. Instead of dithering,~\cite{lee2006two} employs an adaptive directional smoothing filter;~\cite{tu2020adaptive} applies a banding-search algorithm based on distortion visibility. These approaches focus on detection and correction of banding regions, but none enhances the LBD in other areas of a video frame. 

The second approach reconstructs an image’s least-significant bits from the most-significant bits. It became widespread with the arrival of neural networks, but classical algorithms are also worth describing. Many such algorithms are similar to debanding, as they focus on restoring regions that contain false contours. For instance,~\cite{cheng2009bit, wan20122d} propose filling the low bits in accordance with the distance to the false contours. In~\cite{wan2014image, liu2018ipad}, recovery of the low bits employs graph-signal processing and analysis of statistical patterns in a natural image.

Neural-network methods for LSB restoration divide into those that use only one frame and can operate on both images and videos, and those that use multiple frames. The techniques in~\cite{su2019photo, liu2019calf} employ a deep neural-network ResNet to recover missing LSBs. In~\cite{punnappurath2021little}, the authors extend the idea of~\cite{liu2019calf} to reconstruct each subsequent bit plane by training a separate model for each one. The researchers in~\cite{zhang2021acgan} apply a generative adversarial neural network, training the generator to restore the low bits and training the discriminator to determine whether the image is the original HBD or the result of up-conversion. In~\cite{liu2021residual}, the authors propose a multiscale fusion neural network to combine features extracted from the image pyramid.

The work in~\cite{liu2019spatiotemporal, wen20213d, liu2021tanet} uses multiple frames to recover the video’s LSBs. The approach of~\cite{liu2019spatiotemporal} is to align five consecutive frames using motion compensation, then process each frame using ResNet; afterward, the extracted features are combined and fed to ResNet, but inverse convolutions restore the final frame with high color depth. Another variation,~\cite{wen20213d} proposes a neural network with 3D convolutions. The authors of~\cite{liu2021tanet} investigate different ways to align consequent frames for bit-depth reconstruction and conclude that the alignment works best using the global attention module~\cite{wang2018non}. But no existing up-conversion technique can accurately predict the LSBs of a true HBD video; an unnatural appearance caused by bit-depth enhancement is always noticeable on close examination. 
\section{Proposed method}
\begin{figure}[tbp]
\centerline{\includegraphics[width=0.48\textwidth]{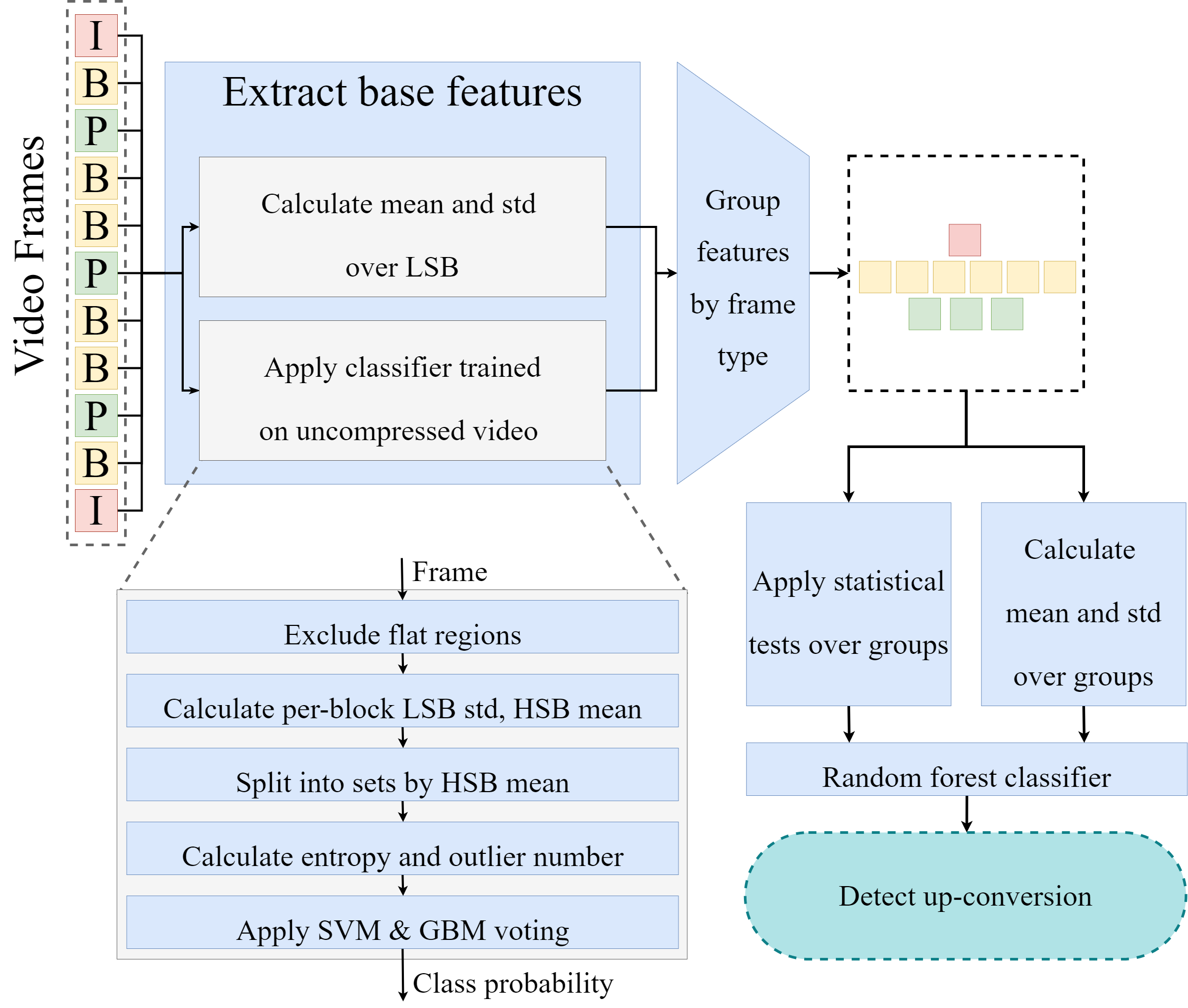}}
\caption{Block diagram of algorithm for detecting bit-depth enhancement.}
\label{fig:main}
\end{figure}

A sequence of compressed video frames containing I, P, and B frames. We trained a classifier to detect whether the dequantization algorithm obtained the two LSBs, and we calculated the following:
\begin{itemize}
\item Average size of compressed I, P, and B frames. This feature is useful, as the stronger the compression, the more the LSBs of the compressed frames differ from those of the original uncompressed frames.
\item Per-frame mean value and standard deviation of the two least-significant bits for each frame type.
\item Statistical tests on simple features to determine for each frame type whether the features belong to a normal distribution and whether the feature values from the given frames belong to the same distribution.
\item Average prediction of a module trained on uncompressed up-converted video for different frame types and statistical tests. We describe this module in detail below.
\end{itemize}
The algorithm extracts the abovementioned features (except the average compressed-frame size) for each color channel of the input video. Block diagram of our algorithm is presented in Figure~\ref{fig:main}. 

\subsection{Detection of bit-depth up-conversion \\ for uncompressed frame}\label{AA}
Detecting bit-depth up-conversion in uncompressed video is easier than doing so after compression, since compression modifies the video’s LSBs and leaves behind flattened traces. For example, checking whether the conversion algorithm has set the LSBs of an uncompressed video to zero is easy. More-complex algorithmic up-conversion necessitates creation of more-sophisticated features. Our proposed method works for a single frame, so it applies even to still images.

Let image $H$ contain the most-significant bits (MSBs) of the color channel $C$ for input frame $I^{d}$ of size $w \times h$, where $d$ is the frame’s bit-depth: \begin{align}C(i,j)=\vec{c}_{i,j} = (b_1, b_2, ... ,b_d)\end{align} 
\begin{align}H(i,j)=(b_1, b_2, ... ,b_{d-2})\end{align} 
The image $L$ contains two LSBs of the input frame $C$: 
\begin{align}L(i,j)=(b_{d-1}, b_d)\end{align} 

The algorithm must determine whether the frame’s two LSBs are original or obtained by up-conversion. We therefore estimate how much the standard deviation $Std(i, j)$ of $L$ depends on the color intensity $Int(i, j)$. We calculate these values in square areas with side length $2s + 1$ and center on each image pixel, except those adjacent to the frame borders. Our estimate of the intensity $Int(i, j)$ is the average of the MSBs in each square region.


Flat areas should be discarded. In addition, we also introduce $D(i, j)$ as standard deviation over $H$.
Flat areas should be discarded. So in addition, we also introduce $D(i, j)$ as the standard deviation over $H$. If $D$ is zero, the MSBs are the same in the area of the pixel at coordinates $i$ and $j$, and the LSBs in this area should be nearly identical. Often, this effect is observable in under- and overexposed image areas. Such distortions can affect visual quality. One goal of HDR video is to reduce the size of these regions, but small highlighted and underexposed areas are common and generally have little effect on visual quality. Therefore, we exclude from consideration the blocks for which $D$ is zero. Thus, we obtain the set of points $P$:

\begin{align}
P = \{(Int(i, j); Std(i, j)): D(i, j) > 0\}
\end{align}

\begin{figure*}[tbp]
\centering{\includegraphics[width=\textwidth]{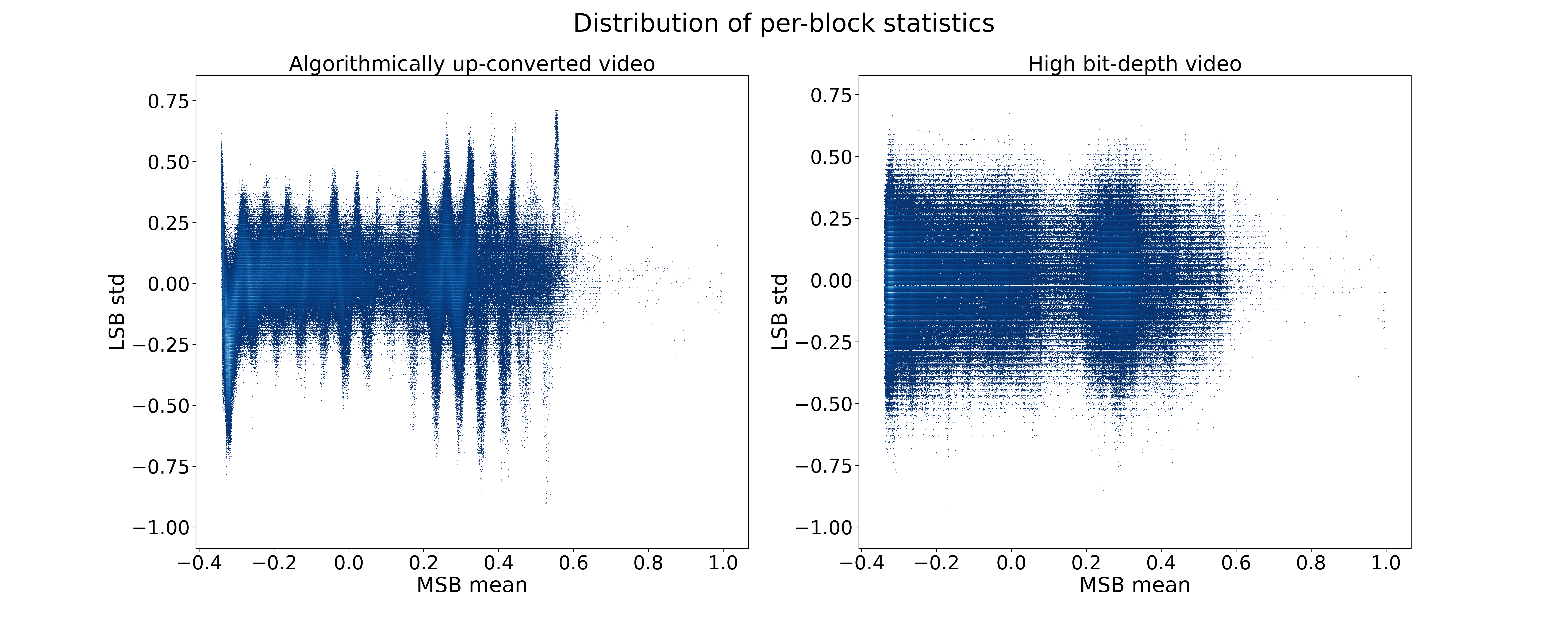}}
\caption{Example $P$ sets for a frame up-converted using ALBM~\cite{punnappurath2021little} and the true HBD frame. At left are the characteristics of the up-converted image; at right, those of the true HBD image.}
\label{fig:cloud}
\end{figure*}

We normalize the data so the mean is zero and the maximum deviation from the mean is less than or equal to one. As the Figure~\ref{fig:cloud} shows, the point set for true HBD images is more uniform along the x-axis than a point set corresponding to the same image with artificially enhanced bit-depth. This effect is observable in most images. We evaluate uniformity along the x-axis by dividing the values into $n$ bins:

\begin{align}
 S_{m} = \{(x, y) \in P: x \in \left[\frac{m-1}{n}k + b, \frac{m}{n}k + b\right]\}
\end{align}
\begin{align}
k=\underset{(x, y) \in P}{\max}x - \underset{(x, y) \in P}{\min}x
\end{align}
\begin{align}
b=\underset{(x, y) \in P}{\min}x
\end{align}

We empirically chose the number of bins to be $n = 100$. Next, our algorithm calculates the mean and standard deviation of each bin. For each mean and standard deviation across all bins, we calculate the number of outliers using the three-sigma rule. The entropy is then the following:
\begin{align}
Entropy=-\sum_{m=1}^{n}\delta_{m}\log\delta_{m}
\end{align}
\begin{align}
\delta_{m} = \underset{(x, y) \in S_{m}}{\max}y - \underset{(x, y) \in S_{m}}{\min}y
\end{align}

The entropy and number of outliers that we calculate for each color channel form the feature vector. For classification, our approach uses voting by two classifiers: support-vector machine (SVM)~\cite{cortes1995support} and gradient-boosting machine (GBM)~\cite{ke2017lightgbm}. Both allow us to estimate a posteriori class-membership probabilities. We employ the average probability as a feature in subsequent bit-depth-enhancement detection for compressed videos. Our algorithm requires just one frame but no additional information about the compressed video’s structure, so it can function on both uncompressed video and still images. But the algorithm deals less well with video frames after compression, which causes features of the original HBD images to become similar to those of up-converted images.

\subsection{Detection of bit-depth up-conversion \\ in compressed video}
To detect bit-depth conversion in compressed video, we propose estimating the correlation of corresponding features for different frame types, which employ different compression techniques and therefore exhibit different degrees of distortion compared with the uncompressed original frames. We calculate basic features: the mean and standard deviation of the LSBs, the average values for each frame type, and the average probability predicted by the SVM and GBM classifiers. For the original HBD videos, we then consider the sets of features for different frame types as samples from the normal distribution. For each set of I-, P-, and B-frames we compute the Shapiro-Wilk~\cite{10.1093/biomet/52.3-4.591} test, examining the null hypothesis that the feature distributions are normal. Our approach then employs the resulting statistics as features in the classifier. Afterward, we use Student’s t-test~\cite{walpole1993probability} to evaluate the hypothesis that the samples from each pair of frame types belong to the same distribution. We calculate the described characteristics for each color channel and apply a random forest classifier to detect bit-depth up-conversion in a compressed video.
\section{Experiments}
\subsection{Datasets}
To train, validate, and test the algorithm, we prepared a dataset based on videos published by CableLabs~\cite{cablelabs}. We refer to this dataset as CableLabs. From the original set we took nine source clips of 3–6 minutes in duration, each having 10 bit-depth. Our next step was to use VQMT~\cite{vqmt} to split the clips into scenes of at least 50 frames. The result was 169 uncompressed videos, forming the basis of the dataset. To obtain up-conversion examples, we replaced the lower two bits with zeros, noise, and the results of ALBM~\cite{punnappurath2021little}. We compressed the resulting videos using the H.265 codec in FFmpeg~\cite{tomar2006converting}, with the CRF (constant rate factor) parameter in the 0–24 range. In total, the result was a dataset of about 8,000 training videos. We applied leave-one-out (LOO) cross-validation~\cite{sammut2010leave} to fine-tune the classifier parameters as well as to estimate classification quality. In LOO, all videos that belong to the same source film were either in the training set or in the validation set. Thus, the model underwent training on the videos obtained from eight clips and then validation on one clip; afterward, we averaged nine results over the clips. 

To test more conversion methods, we collected a dataset of publicly available videos from Vimeo. We refer to this dataset as Vimeo. Our selection included only videos with a low compression rate and 10-bit color depth. From them we chose 16 initial videos and replaced their LSBs with the prediction results of bit-depth enhancement algorithms. Among these algorithms were neural-network bit-depth enhancement methods ALBM~\cite{punnappurath2021little}, BE-CNN~\cite{su2019photo}, BE-CALF~\cite{liu2019calf}, RMFNet~\cite{liu2021residual}, and BitNet~\cite{byun2018bitnet}; classical bit-depth enhancement algorithms CA~\cite{wan20122d}, CRR~\cite{cheng2009bit}, MRC~\cite{mittal2012bit}, and IPAD~\cite{liu2018ipad}; and debanding algorithms AdaDeband~\cite{tu2020adaptive}, deband, and gradfun. FFmpeg implements both deband and gradfun.

\begin{table}[!b]
\caption{Detection quality for uncompressed and compressed videos}
\label{tab:nocompr} \centering
\begin{tabular}{l l | c c}
  \hline
  \multirow{2}{*}{Dataset} & \multirow{2}{*}{Bit-depth} & \multicolumn{2}{c}{F1-score} \\
    &   & Uncompressed & Compressed \\
  \hline
  CableLabs & $6$ to $8$ &$0.9877 \pm 0.0118$ & $0.9564 \pm 0.0080$  \\
  & $8$ to $10$&$0.9605 \pm 0.0361$ & $0.8742 \pm 0.0085$  \\
  Vimeo & $6$ to $8$ & $0.9213$ & $0.9247$ \\
   & $8$ to $10$ & $0.8943$ & $0.8028$ \\
  \hline
\end{tabular}
\end{table}

\subsection{Evaluation on uncompressed videos}
The training process involved only CableLabs. We used LOO cross-validation to estimate an average F1-score. Testing employed 676 videos from this dataset. Vimeo contains videos captured using unknown cameras and converted by unknown algorithms, including slight compression. Testing on this dataset used 224 videos. Table~\ref{tab:nocompr} shows the algorithm-quality evaluation for both datasets.

Vimeo contains outputs from 12 bit-depth enhancement algorithms; only one of these outputs appears in CableLabs, and we investigated it during feature creation and model training. We evaluated the method for each up-conversion algorithm separately. Table~\ref{tab:nocompr-detail} contains results from this comparison. On most up-conversion algorithms, our method shows good quality for both cases: detection of video up-conversion from 6 bit-depth to 8 bit-depth and detection of video up-conversion from 8 bit-depth to 10 bit-depth. This outcome indicates good generalizability of our method, since its training excluded videos converted by these algorithms. 
\begin{table}[!tb]
\caption{Detection quality for various up-conversion algorithms processing uncompressed videos}
\label{tab:nocompr-detail} \centering
\begin{tabular}{l | c c}
  \hline
  \multirow{2}{*}{\makecell{Up-conversion\\algorithm}} & \multicolumn{2}{c}{F1-score} \\
  & 6 to 8 bit & 8 to 10 bit \\
  \hline
  ALBM~\cite{punnappurath2021little} & $0.93$ &  $0.92$\\
  BE-CNN~\cite{su2019photo} &  $0.91$  & $0.91$\\
  BE-CALF~\cite{liu2019calf} &  $0.90$ & $0.90$\\
  RMFNet~\cite{liu2021residual} &  $0.76$   & $0.74$\\
  BitNet~\cite{byun2018bitnet} &  $0.87$   & $0.83$\\
  CA~\cite{wan20122d} &  $0.92$   & $0.90$\\
  CRR~\cite{cheng2009bit} &  $0.90$  & $0.90$\\
  MRC~\cite{mittal2012bit} &  $0.90$  & $0.91$\\
  IPAD~\cite{liu2018ipad} &  $0.90$  & $0.87$\\
  AdaDeband~\cite{tu2020adaptive} &  $0.94$ & $-$\\
  deband &  $0.56$  & $0.52$\\
  gradfun &  $0.80$  & $0.69$ \\
  zeros &  $0.94$  & $0.92$ \\
  noise &  $0.92$  & $0.89$ \\
  \hline
\end{tabular}
\end{table}

\subsection{Evaluation on compressed videos}
Table~\ref{tab:nocompr} shows that the algorithm performs well on both datasets. The detection quality for up-conversion from 6 to 8 bit-depth videos is higher than that from 8 to 10 bit-depth. CableLabs contains about 9,500 videos; Vimeo, about 3,000. 

\begin{figure}[tbp]
\centerline{\includegraphics[width=0.50\textwidth]{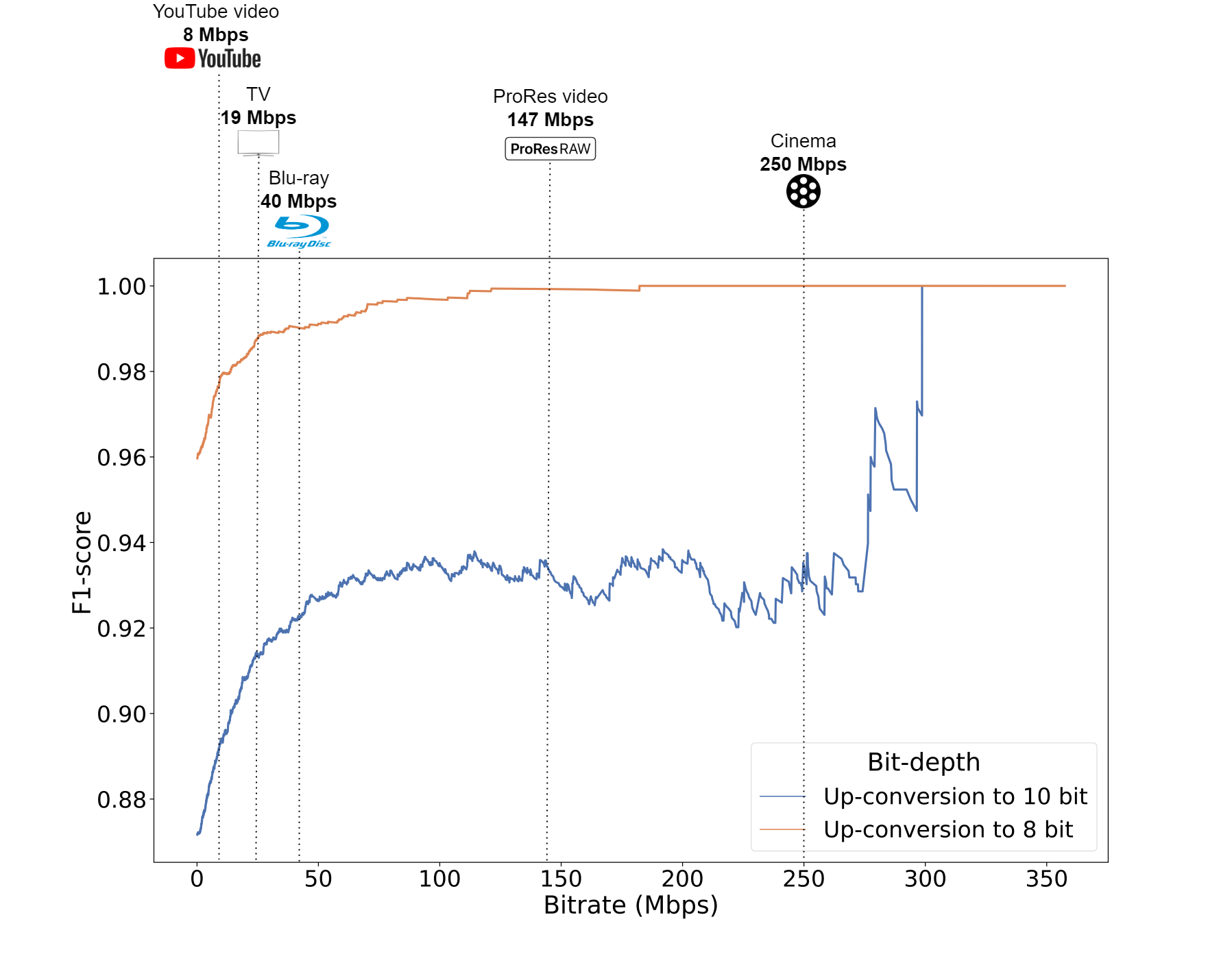}}
\caption{Detection quality of proposed algorithm by F1-score \\ in terms of bitrate.}
\label{fig:bitrate}
\end{figure}

To study the effect of compression ratio on the algorithm’s quality, we plotted the F1-score against bitrate. As Figure~\ref{fig:bitrate} shows, the quality decreases as bitrate decreases, and for low values, the detection quality becomes less reliable. In practical cases, however, the bitrate is higher and the algorithm shows good quality.

\subsection{Ablation study}
\begin{table}[!b]
\caption{Detection quality for full and reduced method}
\label{tab:ablation} \centering
\begin{tabular}{c l | c}
  \hline
  Method & Bit-depth & F1-score \\
  \hline
  \multirow{2}{*}{proposed method} & $6$ to $8$ bit &$0.9247$  \\
  & $8$ to $10$ bit &$0.8028$   \\
  \hline
  \multirow{2}{*}{\makecell{proposed method \\ w\textbackslash o uncompressed image up-conversion detector}}& $6$ to $8$ bit & $0.7256$ \\
   & $8$ to $10$ bit & $0.5110$  \\
  \hline
\end{tabular}
\end{table}

We conducted a study of proposed method for redundancy. Our results show that the method’s quality falls when training occurs without pretraining the SVM and GBM classifiers on uncompressed frames. We tested two variants of the method trained on CableLabs and tested on Vimeo. As Table~\ref{tab:ablation} shows, removal of SVM and GBM reduces quality compared with the full method.

\subsection{Computational Complexity}
Our method works at a frame rate of 15.34 fps on an Intel Core i7-12700K processor running at 3.7GHz. We estimated this processing speed using FullHD videos from the CableLabs dataset.

\section{Conclusion}
In this paper we described a novel approach to bit-depth enhancement detection for compressed video. Proposed method could help analyze a video’s true bit-depth and predict its quality when displayed on a high-contrast screen. It employs statistical analysis of the correlation between different compressed-video frame types to detect bit-depth processing before compression. The technique performed well during extensive experiments and provides good generalization capability for detecting videos produced by unknown up-conversion algorithms. It works well for uncompressed videos as well as for videos compressed using FFmpeg’s H.265 encoder. Modern video codecs, however, can apply different postprocessing, whose effect on our method’s classification quality has yet to undergo testing.

\section{Acknowledgment}
This study received support from the Russian Science Foundation under grant 22-21-00478, \url{https://rscf.ru/en/project/22-21-00478/}.
\bibliography{main.bbl}{}

\begin{thebibliography}{10}
\providecommand{\url}[1]{#1}
\csname url@samestyle\endcsname
\providecommand{\newblock}{\relax}
\providecommand{\bibinfo}[2]{#2}
\providecommand{\BIBentrySTDinterwordspacing}{\spaceskip=0pt\relax}
\providecommand{\BIBentryALTinterwordstretchfactor}{4}
\providecommand{\BIBentryALTinterwordspacing}{\spaceskip=\fontdimen2\font plus
\BIBentryALTinterwordstretchfactor\fontdimen3\font minus
  \fontdimen4\font\relax}
\providecommand{\BIBforeignlanguage}[2]{{%
\expandafter\ifx\csname l@#1\endcsname\relax
\typeout{** WARNING: IEEEtran.bst: No hyphenation pattern has been}%
\typeout{** loaded for the language `#1'. Using the pattern for}%
\typeout{** the default language instead.}%
\else
\language=\csname l@#1\endcsname
\fi
#2}}
\providecommand{\BIBdecl}{\relax}
\BIBdecl

\bibitem{recommendation2012parameter}
I.-R. Recommendation, ``Parameter values for ultra-high definition television
  systems for production and international programme exchange,''
  \emph{International Telecommunication Union, Geneva}, 2012.

\bibitem{Bot2021ASSESSMENTOQ}
E.~L. Bot and T.~Pouli, ``Assessment of quantization artifacts in sdr and hdr
  and cnn-based correction,'' 2021.

\bibitem{bhagavathy2009multiscale}
S.~Bhagavathy, J.~Llach, and J.~Zhai, ``Multiscale probabilistic dithering for
  suppressing contour artifacts in digital images,'' \emph{IEEE Transactions on
  Image Processing}, vol.~18, no.~9, pp. 1936--1945, 2009.

\bibitem{baugh2014advanced}
G.~Baugh, A.~Kokaram, and F.~Piti{\'e}, ``Advanced video debanding,'' in
  \emph{Proceedings of the 11th European Conference on Visual Media
  Production}, 2014, pp. 1--10.

\bibitem{lee2006two}
J.~W. Lee, B.~R. Lim, R.-H. Park, J.~S. Kim, and W.~Ahn, ``Two-stage false
  contour detection algorithm using re-quantization and directional contrast
  features and its application to adaptive false contour reduction,'' in
  \emph{2006 Digest of Technical Papers International Conference on Consumer
  Electronics}.\hskip 1em plus 0.5em minus 0.4em\relax IEEE, 2006, pp.
  377--378.

\bibitem{tu2020adaptive}
Z.~Tu, J.~Lin, Y.~Wang, B.~Adsumilli, and A.~C. Bovik, ``Adaptive debanding
  filter,'' \emph{IEEE Signal Processing Letters}, vol.~27, pp. 1715--1719,
  2020.

\bibitem{cheng2009bit}
C.-H. Cheng, O.~C. Au, C.-H. Liu, and K.-Y. Yip, ``Bit-depth expansion by
  contour region reconstruction,'' in \emph{2009 IEEE International Symposium
  on Circuits and Systems}.\hskip 1em plus 0.5em minus 0.4em\relax IEEE, 2009,
  pp. 944--947.

\bibitem{wan20122d}
P.~Wan, O.~C. Au, K.~Tang, Y.~Guo, and L.~Fang, ``From 2d extrapolation to 1d
  interpolation: Content adaptive image bit-depth expansion,'' in \emph{2012
  IEEE International Conference on Multimedia and Expo}.\hskip 1em plus 0.5em
  minus 0.4em\relax IEEE, 2012, pp. 170--175.

\bibitem{wan2014image}
P.~Wan, G.~Cheung, D.~Florencio, C.~Zhang, and O.~C. Au, ``Image bit-depth
  enhancement via maximum-a-posteriori estimation of graph ac component,'' in
  \emph{2014 IEEE International Conference on Image Processing (ICIP)}.\hskip
  1em plus 0.5em minus 0.4em\relax IEEE, 2014, pp. 4052--4056.

\bibitem{liu2018ipad}
J.~Liu, G.~Zhai, A.~Liu, X.~Yang, X.~Zhao, and C.~W. Chen, ``Ipad: Intensity
  potential for adaptive de-quantization,'' \emph{IEEE Transactions on Image
  Processing}, vol.~27, no.~10, pp. 4860--4872, 2018.

\bibitem{su2019photo}
Y.~Su, W.~Sun, J.~Liu, G.~Zhai, and P.~Jing, ``Photo-realistic image bit-depth
  enhancement via residual transposed convolutional neural network,''
  \emph{Neurocomputing}, vol. 347, pp. 200--211, 2019.

\bibitem{liu2019calf}
J.~Liu, W.~Sun, Y.~Su, P.~Jing, and X.~Yang, ``Be-calf: bit-depth enhancement
  by concatenating all level features of dnn,'' \emph{IEEE Transactions on
  Image Processing}, vol.~28, no.~10, pp. 4926--4940, 2019.

\bibitem{punnappurath2021little}
A.~Punnappurath and M.~S. Brown, ``A little bit more: Bitplane-wise bit-depth
  recovery,'' \emph{IEEE Transactions on Pattern Analysis and Machine
  Intelligence}, 2021.

\bibitem{zhang2021acgan}
J.~Zhang, Q.~Dou, J.~Liu, Y.~Su, and W.~Sun, ``Be-acgan: Photo-realistic
  residual bit-depth enhancement by advanced conditional gan,''
  \emph{Displays}, vol.~69, p. 102040, 2021.

\bibitem{liu2021residual}
J.~Liu, X.~Wen, W.~Nie, Y.~Su, P.~Jing, and X.~Yang, ``Residual-guided
  multiscale fusion network for bit-depth enhancement,'' \emph{IEEE
  Transactions on Circuits and Systems for Video Technology}, 2021.

\bibitem{liu2019spatiotemporal}
J.~Liu, P.~Liu, Y.~Su, P.~Jing, and X.~Yang, ``Spatiotemporal symmetric
  convolutional neural network for video bit-depth enhancement,'' \emph{IEEE
  Transactions on Multimedia}, vol.~21, no.~9, pp. 2397--2406, 2019.

\bibitem{wen20213d}
G.~Wen, Z.~Yang, C.~Li, R.~Xie, L.~Song, and W.~Cai, ``3d-bitnet: Flow-agnostic
  and precise network for video bit-depth expansion,'' in \emph{2021 IEEE
  International Symposium on Broadband Multimedia Systems and Broadcasting
  (BMSB)}.\hskip 1em plus 0.5em minus 0.4em\relax IEEE, 2021, pp. 1--6.

\bibitem{liu2021tanet}
J.~Liu, Z.~Yang, Y.~Su, and X.~Yang, ``Tanet: Target attention network for
  video bit-depth enhancement,'' \emph{IEEE Transactions on Multimedia}, 2021.

\bibitem{wang2018non}
X.~Wang, R.~Girshick, A.~Gupta, and K.~He, ``Non-local neural networks,'' in
  \emph{Proceedings of the IEEE conference on computer vision and pattern
  recognition}, 2018, pp. 7794--7803.

\bibitem{cortes1995support}
C.~Cortes and V.~Vapnik, ``Support-vector networks,'' \emph{Machine learning},
  vol.~20, no.~3, pp. 273--297, 1995.

\bibitem{ke2017lightgbm}
G.~Ke, Q.~Meng, T.~Finley, T.~Wang, W.~Chen, W.~Ma, Q.~Ye, and T.-Y. Liu,
  ``Lightgbm: A highly efficient gradient boosting decision tree,''
  \emph{Advances in neural information processing systems}, vol.~30, 2017.

\bibitem{10.1093/biomet/52.3-4.591}
S.~S. Shapiro and M.~B. Wilk, ``{An analysis of variance test for normality
  (complete samples)},'' \emph{Biometrika}, vol.~52, no. 3-4, pp. 591--611,
  1965.

\bibitem{walpole1993probability}
R.~E. Walpole, R.~H. Myers, S.~L. Myers, and K.~Ye, \emph{Probability and
  statistics for engineers and scientists}.\hskip 1em plus 0.5em minus
  0.4em\relax Macmillan New York, 1993, vol.~5.

\bibitem{cablelabs}
\BIBentryALTinterwordspacing
{CableLabs UHD Video Dataset}. [Online]. Available:
  \url{https://www.cablelabs.com/4k}
\BIBentrySTDinterwordspacing

\bibitem{vqmt}
\BIBentryALTinterwordspacing
{MSU Video Quality Measurement Tool}. [Online]. Available:
  \url{https://compression.ru/video/quality_measure/vqmt_download.html}
\BIBentrySTDinterwordspacing

\bibitem{tomar2006converting}
S.~Tomar, ``Converting video formats with ffmpeg,'' \emph{Linux Journal}, vol.
  2006, no. 146, p.~10, 2006.

\bibitem{sammut2010leave}
C.~Sammut and G.~I. Webb, ``Leave-one-out cross-validation,''
  \emph{Encyclopedia of machine learning}, pp. 600--601, 2010.

\bibitem{byun2018bitnet}
J.~Byun, K.~Shim, and C.~Kim, ``Bitnet: Learning-based bit-depth expansion,''
  in \emph{Asian Conference on Computer Vision}.\hskip 1em plus 0.5em minus
  0.4em\relax Springer, 2018, pp. 67--82.

\bibitem{mittal2012bit}
G.~Mittal, V.~Jakhetiya, S.~P. Jaiswal, O.~C. Au, A.~K. Tiwari, and D.~Wei,
  ``Bit-depth expansion using minimum risk based classification.'' in
  \emph{2012 IEEE Visual Communications and Image Processing (VCIP)}, 2012, pp.
  1--5.

\end{thebibliography}
\bibliographystyle{IEEEtran}

\end{document}